\documentclass{article}

\usepackage{microtype}
\usepackage{graphicx}
\usepackage{subfigure}
\usepackage{booktabs} %

\usepackage{hyperref}
\usepackage[nameinlink]{cleveref}

\crefname{chapter}{Chapter}{Chapters}
\crefname{section}{Section}{Sections}
\crefname{subsection}{Section}{Sections}
\crefname{subsubsection}{Section}{Sections}
\crefname{figure}{Figure}{Figures}
\crefname{table}{Table}{Tables}
\crefname{subfigure}{Figure}{Figures}
\crefname{page}{Page}{Pages}
\crefname{equation}{Equation}{Equations}
\crefname{appendix}{Appendix}{Appendix}

\usepackage[accepted]{icml2019}

\icmltitlerunning{A Perspective on Objects and Systematic Generalization in Model-Based RL} %

\begin{document}

\twocolumn[
\icmltitle{A Perspective on Objects and Systematic Generalization in Model-Based RL} %

\icmlsetsymbol{equal}{*}

\begin{icmlauthorlist}
\icmlauthor{Sjoerd van Steenkiste*}{idsia}
\icmlauthor{Klaus Greff*}{idsia}
\icmlauthor{J\"{u}rgen Schmidhuber}{idsia,nnaisense}
\end{icmlauthorlist}

\icmlaffiliation{idsia}{The Swiss AI Lab - IDSIA, USI \& SUPSI}
\icmlaffiliation{nnaisense}{NNAISENSE}

\icmlcorrespondingauthor{Sjoerd van Steenkiste}{sjoerd@idsia.ch}

\icmlkeywords{Objects, Systematic Generalization, Model-based RL} %

\vskip 0.3in
]

\printAffiliationsAndNotice{\icmlEqualContribution} %

\begin{abstract}

In order to meet the diverse challenges in solving many real-world problems, an intelligent agent has to be able to dynamically construct a model of its environment.
Objects facilitate the modular reuse of prior knowledge and the combinatorial construction of such models. %
In this work, we argue that dynamically bound features (objects) do not simply emerge in connectionist models of the world.
We identify several requirements that need to be fulfilled in overcoming this limitation and highlight corresponding inductive biases. %

\end{abstract}

\section{Introduction}
\label{sec:introduction}

Artificial General Intelligence (AGI) requires an intelligent agent to solve a wide variety of real-world tasks.
Learning to solve these tasks efficiently involves sharing knowledge between tasks, and systematic generalization from relatively few samples. 
In contrast, agents trained with Reinforcement Learning (RL) frequently fall short in this regard: they rely on excessive amounts of data~\cite{francois-lavet2018introduction} and are unable to generalize beyond their initial training regime~\cite{leike2017ai}. %

Model-based RL promises to alleviate this problem by using a general (non task-specific) world model that captures the latent structure of the environment.
This more abstract knowledge about the world is expected to be useful for many (even novel) tasks and facilitate simulation and planning.
It is unclear what constitutes a good model, and frequently models are either engineered~\cite{deavilabelbute-peres2018endtoend} or obtained by training a deep (recurrent) neural network to predict future states of the world~\cite{schmidhuber1990making,ha2018recurrent}. 
In the latter case, an underlying assumption is that the learned representations of such a network present suitable abstractions for transfer and planning, analogous to the versatility of features learned by a deep convolutional image classifier.
However, the limited success of learned models for model-based RL in these domains raises doubts about the validity of this assumption.

In this work, we argue that, rather than learning a single monolithic model that handles all situations in all environments, what is needed is a flexible system which dynamically infers a suitable model on the fly. %
A human playing the game of \emph{Space Invaders} uses a mental model that revolves around space ships and aliens, without simultaneously also considering all other aspects of the real world that are relevant for other tasks.
Humans are also quick to adapt their model to new information by adding or removing additional assumptions.
For example, reading the manual of a game before playing greatly increases first-episode performance~\cite{tsividis2017human}. %

Initially, it may appear that in arguing for a dynamic model we have mostly made the task of model-learning harder: we now require learning many different models that fit specific situations. 
Why then would we expect such a model to perform any better or even work at all?

\paragraph{Objects}
Objects are the key piece to this puzzle, in that they facilitate the modular reuse of prior knowledge and the combinatorial construction of novel models.
It is well-established that objects play a central role in human cognition, both for internal reasoning and as the basis for communicating about the world.
Indeed, objects are widely considered to be core knowledge~\cite{spelke2007core}, and infants learn about objects already within their first year of life~\cite{munakata1997rethinking}.

RL methods that leverage the combinatorics of objects and relations have shown similar benefits in terms of systematic generalization~\cite{zambaldi2019deep}, sample efficiency~\cite{diuk2008objectoriented}, and transferring skills and knowledge across domains~\cite{kansky2017schema}. %
Recently, there appears to be an emerging consensus that objects are important in learning intelligent agents~\cite{lake2017building}, while it remains unclear in how to fully realize this potential.
The discrete and compositional nature of objects seems at odds with many of the core tenets of connectionism, and they are unlikely to emerge naturally in neural networks.
Reconciling the two is a difficult problem and requires careful thought to ensure a synergistic integration. %

\paragraph{The Binding Problem}

How then should we think about objects?
Why do they not simply emerge in neural networks, what is missing, and how can this be addressed?
Many of these questions have been raised and debated in theoretical neuroscience and have become known as the binding problem: How does the brain bind features together into objects while keeping them separate from other objects.
Inspired by this literature (cf. \citet{treisman1999solutions, vondermalsburg1995binding}) we will focus on three main challenges in incorporating objects in connectionist models of the world: segregation, representation, and composing, which we discuss in the next sections.

Segregation is about object discovery, i.e. given a set of observations, what are good candidates for representational objects and how can they be extracted.
Representation is about storing this representational content in neural networks, and as we will find, plain fully-connected feedforward networks are ill-equipped to solve this task.
Finally, composition is about using representational objects efficiently in a way that ensures combinatorial generalization (\emph{systematicity}; \citet{niklasson1994being}).

\section{Representation}
\label{sec:representation}

What are good object representations? 
If objects are to serve as the primitives for compositional reasoning, it is important that their representations support that end.
Here we argue for three main requirements:\footnote{Keeping space limitations in mind we will not attempt at exhaustively listing all requirements. Instead we focus on those that we believe to be most important.}

\begin{description}
  \item[Universal] Each object representation should be able to represent any object regardless of position, class or other properties. 
  It should facilitate generalization, even to unseen objects (zero-shot generalization), which in practice means that its representation should be distributed and disentangled.
  \item[Multi-Object] It should be possible to represent multiple objects simultaneously, such that they can be related and composed but also transformed individually. 
  This only needs to cover a small number of objects at the same time (e.g. $7\pm2$; \citealt{miller1956magical}), since there is an intractable number of possible objects. 
  Instead, objects should be swapped in and out of this working memory on demand.
  \item[Common Format] All objects should be represented in the same format, i.e. in terms of the same features. 
  This makes representations comparable, provides a unified interface for compositional reasoning and allows the transfer of knowledge between objects. %
\end{description}

It is easy to see how regular representations of fully-connected neural networks fall short in this regard:
When representing multiple objects, they can either reuse the same features for all objects simultaneously, thus superimposing representations which leads to ambiguities (\{red, square\} + \{blue, triangle\} = \{red, blue, square, triangle\}).
Alternatively, they can allocate a different set of features per object which violates common format.
Without any architectural bias in the form of weight sharing, useful multi-object representations are thus unlikely to emerge naturally in a neural network.
In what way can this problem be addressed?

\begin{figure}
\centering
\includegraphics[width=\linewidth]{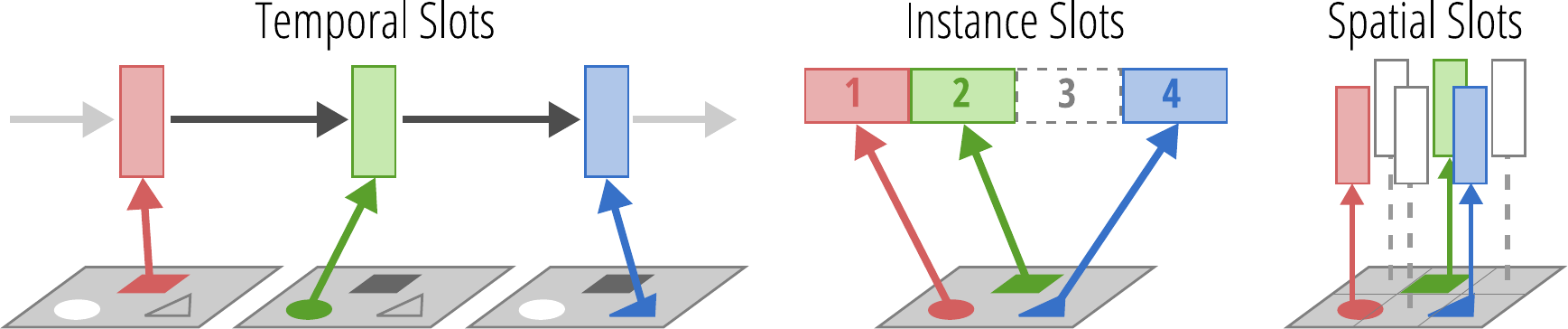}
\caption{Different types of \emph{slot-based} representation strategies.}
\label{fig:slots}
\vspace{-10pt}
\end{figure}

Weight sharing, as it is used, for example, in ConvNets and RNNs, is a step in the right direction.
We call these approaches ``slot-based'' because they provide several slots that all share weights and can thus be used to represent objects in a common format.
In the case of RNNs there is one slot per time-step~\cite{eslami2016attend}, while in ConvNets there is one slot per spatial location in the image~\cite{santoro2017simple}.
Note that both are in slight violation of universality because they tie a slot to a specific time step or location, while RNNs additionally do not \emph{simultaneously} represent multiple objects. 
We can extend the idea of representational slots and consider a setting in which each object has its own universal slot and all slots share a common format (instance slots, c.f.~\cref{fig:slots}).
While this constitutes a good object representation, it raises another problem: if all slots are identical and share weights, then how do they not end up all representing the same object?
Solving this conundrum requires a dynamic information routing process that goes beyond simple feed-forward processing (see~\cref{sec:segregation}). 

There are two, less developed, alternatives to slot-based approaches that have the potential to meet our requirements: 
\emph{Augmentation} approaches keep a single set of features but augment each feature to include some extra grouping information. 
Examples include complex-valued activations (e.g. \citet{reichert2013neuronal}) or spiking networks that encode grouping via synchronization (e.g. \citet{lane1998simple}).
\emph{Embedding} approaches carefully embed multi-object representations in a higher-dimensional space (e.g. \emph{Tensor Product Representations};~\citealt{smolensky1990tensor}).

\section{Segregation}
\label{sec:segregation}

It can be difficult to provide precise boundaries or definitions even for concrete objects like a tree, a mountain, and a river.
Matters become even worse with slightly more abstract objects like a hole, a shadow, or a street corner.
Clearly, sensory information does not come pre-structured into objects, yet we so effortlessly and consistently perceive them. %
How can we aid our agents in developing an equally general understanding of objects?
We address this by focusing on the role of objects as computational primitives in a compositional reasoning system, namely as abstract patterns of the input data that are modular, dynamic, and consistent:

\begin{description}
  \item[Modular] Objects subdivide the input into parts with strong internal coherence while being mostly independent of each other given some task under consideration.
  This division can be thought of as a form of clustering by mutual predictability and helps minimize the error that results from treating them as independent entities.

  \item[Dynamic] Objects are task-dependent, i.e. there is no one fixed definition of objects that applies to all tasks.
  For example, objects can be part-whole hierarchies whose parts are objects themselves: a stack of chairs can be viewed as a single object (the stack) or as multiple objects (the individual chairs). 
  It necessitates top-down feedback: interaction between the up-stream problem solving and down-stream segregation to obtain a dynamic definition of objects.
  \item[Consistent] 
  Representational objects often ``refer to'' physical objects in the real world (although this does not need to be the case), and their usefulness depends on the reliability of that link. 
  The output of the segregation process must thus be stable and consistent to ensure that the results from internal reasoning can be mapped back onto the environment.
  Consistency is also important in communication (different agents should agree on objects), and in the absence of information, e.g. as a result of occlusion.
\end{description}

Modularity rules out standard convolutional neural networks as a means to learn object representations given by the representational content at each spatial slot.
Each convolutional layer with a kernel size exceeding $1 \times 1$ creates dependencies between local spatial neighborhoods. 
Through depth, the representational content of upper layers encode information from all spatial positions and are no longer modular: a change affecting a single object in the input image affects the representations at \emph{all} spatial locations in the upper layers. 

Dynamicity implies that we can not treat segregation as a pre-processing step that extracts objects from input data.
This rules out the use of large quantities of labeled data to pre-train an image segmenter, or the use of domain-specific engineering as is commonly found in generative models that essentially encode a fixed definition of object.
Moreover, human labor is an expensive resource that we can not spend exhaustively to overcome all possible situations.

We conclude that to a large extend object learning must be \emph{unsupervised} through a specialized mechanism that allows for the possibility to incorporate top-down feedback.
Two promising approaches from the literature are \emph{attention} and \emph{differentiable clustering}.

Attention mechanisms are used to selectively attend to a subset of the image, i.e. parts that correspond to a single object~\cite{schmidhuber1991learning, eslami2016attend}.
In this way, attention restricts the information intake and ensures that the resulting representations are modular. 
Top-down feedback can be incorporated by granting control of the attention window to the agent that learns to solve some task~\cite{mnih2014recurrent}.
A downside is that objects are processed in an iterative fashion, which may make it more challenging to reason about multiple objects simultaneously~\cite{kosiorek2018sequential}.

An alternative mechanism is differentiable clustering, which seeks to partition the input in a number of segments while learning the similarity function.
Individual segments are disjoint and result in modularity, while the iterative nature of these clustering procedures allow top-down feedback to be incorporated~\cite{greff2017neural, greff2019multiobject}.

\section{Composition}
\label{sec:composition}

Let us now assume that representation and segregation have been addressed, and we have available a set of relevant independent objects represented in a common format.
Note that when used correctly, these object representations can already make tasks like performing basic feature comparisons very easy.
For example, a function that receives a pair of objects as input and compares their size-related features could easily be learned, and would almost automatically generalize to arbitrary pairs of objects.%

In contrast, combinatorial generalization is not a given for more complex relational reasoning. 
While it also involves learning general functions that accept objects as their arguments, one has to take extra care in being able to flexibly assign the right objects to their corresponding function arguments, as well as in learning about different structural forms that imply different ways of generalizing~\cite{kemp2008discovery}.
These then imply the following requirements:
\begin{description}
\item[General Relations]
Relations differ both in their meaning and in the patterns of generalization that they imply. 
A general reasoning system, therefore, has to be able to instantiate many different types of relations, which necessitates a general representational form.

\item[Dynamic Binding] 
In order to construct a model for a specific situation, the system needs the flexibility to freely combine objects and relations into an arbitrary structure.
Both the structure of relations and the associated objects (\emph{variable binding}; \citealt{browne2000connectionist}) have to be inferred dynamically during run-time.

\item[Role-filler Independence] The content of objects should be independent of their structural roles~\cite{hummel2003symbolicconnectionist}.
That is, any object can take part in any relation, and the interpretation of the whole is determined by both the parts and the structure.
This is related to \emph{common format} and is the key to compositionality that enables the powerful systematic generalization that is characteristic of many symbolic systems.

\end{description}

One approach is to implement complex relational reasoning in a sequential fashion. 
At each step, an object associated with a particular role is processed and the resulting intermediate computation is stored, to be combined in the next step.
While it is clear that a plain RNN can perform this type of computation, the dual role of intermediate representations in representing objects and intermediate computation suggests a very specific function that may be hard to learn~\cite{graves2014neural}.
Alternatively, by combining an RNN with a suitable memory mechanism (eg.~\citet{Das:92,mozer1993connectionist,reed2015neural,graves2016hybrid}) or fast
weights (eg.~\citet{Schmidhuber:92ncfastweights,Schmidhuber:93ratioicann,schlag2018learning}) it may be more easy to learn general functions of this kind.

An alternative approach is to embed objects, and intermediate representations as nodes in a (directed) graph and let computation take place along its edges.
These computation graphs can implement arbitrary relationships, including recursive computation by re-applying the same function successively.
Graph Networks~\cite{battaglia2018relational} structure neural network computations according to this underlying graph and perform relational reasoning through repeated message-passing between the nodes in the graph. %
Compositionality is achieved through weight-sharing, i.e. by learning a general function that operates on (pairs of) nodes following their topological relationship.
However, while graph networks have been successfully applied in the domain of physical reasoning (e.g. \citet{battaglia2016interaction, vansteenkiste2018relational}), a remaining challenge is in dynamically inferring the right structure (i.e. dynamic binding).

While graph networks appear most promising in addressing the challenges of composition, one other approach deserves a mention.
\emph{Embedding approaches}, such as Poincar\'{e} embeddings~\cite{nickel2017poincare}, generalize Euclidean representations to other spaces that more suited in modeling certain types of relations, in this case: hierarchical relationships.
However, the feature representations are essentially adapted to reflect the underlying relation during training, which implies fixed roles and binding during inference.

\section{Conclusion}
\label{sec:conclusion}
We have argued that feature representations alone are inadequate abstractions for planning, reasoning, and for systematically transferring knowledge to novel situations.
To meet the diverse challenges on our quest towards AGI an agent needs to be able to dynamically construct new models about its environment on the fly while reusing as much prior knowledge as possible. 
We have argued that objects (dynamically bound features) are adequate building blocks to quickly and flexibly compose such task-specific models.
Although our examples were centered around vision, we believe that the notion of objects applies equally to other domains like audio, tactile and even abstract thought.
By focusing on their role as compositional primitives, we have identified some inductive biases that we believe are necessary for objects to arise within a connectionist system.
They can be categorized into three areas: representation, segregation, and composition of objects.

Among these three we find that segregation is most frequently neglected and deserves more attention.
Common approaches rely either on some combination of pre-processing pipelines, supervision, or highly engineered generative models of objects.
Meanwhile, the few approaches that tackle this challenge in a holistic and unsupervised way are brittle and have not yet been scaled to real-world data.
Developing better methods for tackling the segregation problem within the framework of connectionism is going to be a central challenge on the way towards AGI.
Similarly, we would like to stress the importance of integrating solutions to all three aspects into a single system.
The potential of objects as modular building blocks can only be realized in full if they are both informed by learned representations, and by feedback from the composite model. 

Another important direction is the integration of objects with other critical cognitive mechanisms such as attention and memory.
Because objects are optimized to be modular, they naturally aggregate features that need to be processed together, but which can be separated from other information.
This makes them ideal primitives for attention and for storage and retrieval from long-term memory.
Attention, in turn, can simplify a task by filtering out irrelevant information and can guide the processing required for more complex reasoning chains.
Such a reasoning process could then also query objects from memory on demand to be compared to or integrated with the current model. 

With this short essay, we hope to draw attention to the intricacies of objects and inspire others to think critically about their integration in connectionist models.

\section*{Acknowledgements}

This research was funded by SNF grant 200021\_165675/1.

\bibliography{references}
\bibliographystyle{icml2019}

\end{document}